\documentclass[10pt]{article}

\usepackage{amsmath}
\usepackage{amsfonts}
\usepackage{mathrsfs}
\usepackage{amssymb}
\usepackage{amsthm}
\usepackage{graphicx}
\usepackage{algorithm}
\usepackage{setspace}

\newtheorem{definition}{Definition}
\newtheorem{proposition}{Proposition}
\newtheorem{lemma}{Lemma}
\newtheorem{theorem}{Theorem}

\begin{document}
\title{Clustering with Transitive Distance and K-Means~Duality}
\author{Chunjing Xu{${}^{\dagger}$}, Jianzhuang Liu{${}^{\dagger}$}, Xiaoou
Tang{${}^{\ddagger}$}\\
{${}^{\dagger}$}{\small \emph{Department of Information Engineering, The
Chinese University of Hong Kong.}}\\
{\small \emph{E-mail: cjxu6,jzliu@ie.cuhk.edu.hk}}\\
{${}^{\ddagger}$}{\small \emph{Microsoft Research Asia, Beijing.}}
{\small \emph{E-mail: xitang@microsoft.com}}
}
\maketitle

\begin{abstract}
Recent spectral clustering methods are a propular and powerful technique for
data clustering. These methods need to solve the eigenproblem whose
computational complexity is $O(n^3)$, where $n$ is the number of data samples.
In this paper, a non-eigenproblem based clustering method is proposed to deal
with the clustering problem. Its performance is comparable to the spectral
clustering algorithms but it is more efficient with computational complexity
$O(n^2)$. We show that with a transitive distance and an observed property,
called K-means duality, our algorithm can be used to handle data sets with
complex cluster shapes, multi-scale clusters, and noise. Moreover, no parameters
except the number of clusters need to be set in our algorithm. 
\end{abstract}

\bigskip
\textbf{Index Term} -- Clustering, duality, transitive distance, ultra-metric.
\bigskip

\section{Introduction}

\label{introsec} Data clustering is an important technique in many applications
such as data mining, image processing, pattern recognition, and
computer vision. Much effort has been devoted to this research~\cite{mac67},
\cite{jain99data}, \cite{shi2000nca}, \cite{ng2001sca}, \cite{girolami2002mkb},
\cite{comaniciu2002msr}, \cite{zelnikmanor2004sts}, \cite{boykov2004ecm}. A
basic principle (assumption) that guides the design of a clustering
algorithm is:

\vspace{0.2cm}
\noindent\textbf{Consistency}: \emph{Data within the same cluster are closed to 
each other, while data belonging to different clusters are relatively far away}.
\vspace{0.2cm}

According to this principle, the hierarchy approach \cite{johnson1967hcs} begins
with a trivial clustering scheme where every sample is a cluster, and then
iteratively finds the closest (most similar) pairs of clusters and merges them
into larger clusters. This technique totally depends on local structure of data,
without optimizing a global function. An easily observed disadvantage of this
approach is that it often fails when a data set consists of multi-scale
clusters \cite{zelnikmanor2004sts}.

Besides the above consistency assumption, methods like the K-means and EM also
assume that a data set has some kind of underlying structures
(hyperellipsoid-shaped or Gaussian distribution) and thus any two clusters can
be separated by hyperplanes. In this case, the commonly-used Euclidean distance
is suitable for the clustering purpose.

With the introduction of kernels, many recent methods like spectral
clustering \cite{ng2001sca}, \cite{zelnikmanor2004sts} consider that clusters in
a
data set may have more complex shapes other than compact sample clouds. In
this general case, kernel-based techniques are used to achieve a reasonable
distance measure among the samples. In \cite{ng2001sca}, the eigenvectors
of the distance matrix play a key role in clustering. To overcome the
problems such as multi-scale clusters in \cite{ng2001sca},
Zelnik-manor and Perona proposed self-tuning spectral clustering, in which the
local scale of the data and the structure of the eigenvectors of the distance
matrix are considered~\cite{zelnikmanor2004sts}. Impressive results have been
demonstrated by spectral clustering and it is regarded as the most promising
clustering technique \cite{verma2003csc}. However, most of the current kernel
related clustering methods, including spectral clustering that is unified to the
kernel K-means framework in \cite{dhillon2004kkm}, need to solve the
eigenproblem, suffering from high computational cost when the data set is
large.

In this paper, we tackle the clustering problem where the clusters can be of
complex shapes. By using a transitive distance measure and an observed property,
called K-means duality, we show that if the consistency condition is satisfied,
the clusters of arbitrary shapes can be mapped to a new space where the clusters
are more compact and easier to be clustered by the K-means algorithm. With
comparable performance to the spectral algorithms, our algorithm does not need
to solve the eigenproblem and is more efficient with computational complexity
$O(n^2)$ than the spectral algorithms whose complexities are $O(n^3)$, where $n$
is the number of samples in a data set.

The rest of this paper is structured as follows. In Section~\ref{distance_sec},
we discuss the transitive distance measure through a graph model of a data
set. In Section~\ref{dualitysec}, the duality of the K-means algorithm is
proposed
and its application to our clustering algorithm is explained.
Section~\ref{secalgo} describes our algorithm and presents a scheme to reduce
the computational complexity. Section~\ref{experiments_sec} shows experimental
results on some synthetic data sets and benchmark data sets, together with
comparisons to the K-means algorithm and the spectral algorithms in
\cite{ng2001sca} and \cite{zelnikmanor2004sts}. The conclusions are given in
Section~\ref{conclusion_sec}.

\section{Ultra-metric and Transitive Distance}
\label{distance_sec}
In this section, we first introduce the concept of ultra-metric and then define
one, called transitive distance, for our clustering algorithm.

\subsection{Ultra-metric}
An ultra-metric $D$ for a set of data samples $V=\{x_i |
i=1,2,\cdots,n\}\subset R^l$ is defined as follows:
\begin{itemize}
\item[1)] $D: V\times V\rightarrow R$ is a mapping, where $R$ is
the set of real numbers.
\item[2)] $D(x_i,x_j)\geq 0$,
\item[3)] $D(x_i,x_j) = 0$ if and only if $x_i=x_j$,
\item[4)] $D(x_i,x_j) = D(x_j,x_i)$,
\item[5)] $D(x_i,x_j) \leq \max\{D(x_i,x_k),D(x_k,x_j)\}$ for any $x_i$, $x_j$,
and $x_k$ in $V$.
\end{itemize}

The last condition is called the \emph{ultra-metric inequality}. The
ultra-metric may seem strange at the first glance, but it appears naturally in
many applications, such as in semantics~\cite{debakker1982dsc} and phylogenetic
tree analysis~\cite{page1998mep}. To have a better understanding of it, we next
show how to obtain an ultra-metric from a traditional metric where the triangle
inequality holds. 

\begin{figure*}[t]
    \centering
    \includegraphics[width=12cm]{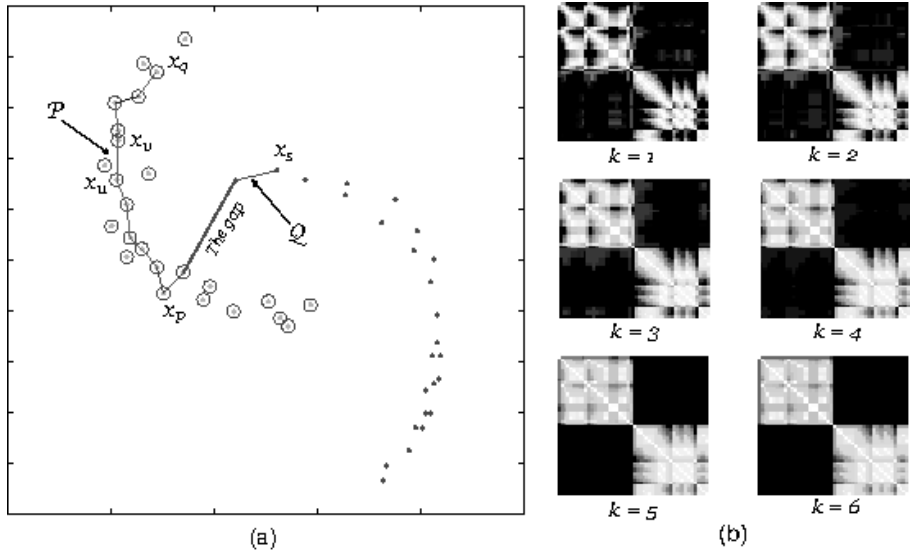}
    \caption{(a) A two-moon data set used to demonstrate the transitive
distance, where samples of one cluster are denoted by circles and samples of
another cluster are denoted by dots. (b) Maps of transitive distance matrices
with different orders. }
\label{diffdistance_fig}
\end{figure*}

In Fig.~\ref{diffdistance_fig},
the distance between samples $x_p$ and $x_q$ is larger than that between $x_p$
and $x_s$ from the usual viewpoint of the Euclidean metric. A more reasonable
metric on the data set should give a closer relationship (thus smaller
distance) between $x_p$ and $x_q$ than that between $x_p$ and $x_s$ since $x_p$
and $x_q$ lie in the same cluster but $x_p$ and $x_s$ do not. A common method
to overcome this difficulty is to create a non-linear mapping
\begin{equation}
\label{mapfund}
\phi: V\subset R^l\rightarrow V'\subset R^s,
\end{equation} such that the
images of any two clusters in $R^s$ can be split linearly. This method is
called the kernel trick and is overwhelmingly used in recent clustering schemes.
Usually the mapping that can reach this goal is hard to find. Besides, another
problem arises when the size of the data set increases; these schemes usually
depend on the solution to the eigenproblem, the time complexity of which is
$O(n^3)$ generally.

Can we have a method that can overcome the above two problems and still achieve
the kernel effect? In Fig.~\ref{diffdistance_fig}(a), we observe that
$x_p$ and $x_q$ are in the same cluster only because the other samples marked by
a circle exist; otherwise it makes no sense to argue that $x_p$ and $x_q$ are
closer than $x_p$ and $x_s$. In other words, the samples marked by a circle
contribute the information to support this observation.

Let us also call each sample a \emph{messenger}. Take $x_u$ as an
example. It brings some messsage from $x_p$ to $x_q$ and vice versa. The way
that $x_p$ and $x_q$ are closer than the Euclidean distance between them can be
formulated as
\begin{equation}
D(x_p, x_q) \leq \max\{d(x_p,x_u),d(x_u,x_q)\},
\label{equbridge}
\end{equation}
where $d(\cdot,\cdot)$ is the Euclidean distance between two samples, and
$D(\cdot,\cdot)$ is the distance we are trying to find that can reflect the true
relationship between samples. In~(\ref{equbridge}), $x_u$ builds a bridge
between $x_p$ and $x_q$ in this formulation. When more and more messengers come
in, we can define a distance through $k$ of these messengers. Let
$\mathcal{P}=x_{u_1}x_{u_2}\cdots x_{u_k}$ be a path with $k$ vertices, where
$x_{u_1}=x_p$ and $x_{u_k}=x_q$. A distance between $x_p$ and $x_q$ with
$\mathcal{P}$ is defined as
\begin{equation}
D_{\mathcal{P}}(x_p,x_q) =
\max_{x_{u_i}x_{u_{i+1}}\in\mathcal{P} \atop 1\leq
i\leq k-1}\{d(x_{u_i},x_{u_{i+1}})\}.
\end{equation}

We show an example in Fig.~\ref{diffdistance_fig}(a), where a path
$\mathcal{P}$ from $x_p$ to
$x_q$ is given. The new distance between $x_p$ and $x_q$
through $\mathcal{P}$ equals $d(x_u,x_v)$, which is smaller than the original
distance $d(x_p,x_q)$. For samples $x_p$ and $x_s$, there are also paths between
them, such as the path $\mathcal{Q}$, which also result in new distances
between them smaller than $d(x_p,x_s)$. However, no matter how the path
is chosen, the new distance between $x_p$ and $x_s$ is always larger than or
equal to the smallest gap between the two clusters as follows.

Given two samples in a data set, we can have many paths connecting them.
Therefore we define the new distance, called the \emph{transitive
distance}, between two samples as follows.

\begin{definition}\label{transdk}
Given the Euclidean distance $d(\cdot,\cdot)$, the derived transitive distance
between
samples $x_p,x_q\in V$ with order $k$ is defined as
\begin{equation}
D_k(x_p,x_q) =
\min_{\mathcal{P}\in\mathbb{P}_k}\max_{e\in\mathcal{P}}\{d(e)\},
\label{transdis_equ}
\end{equation}
where $\mathbb{P}_k$ is the set of paths connecting $x_p$ and $x_q$, each such
path is composed of at most $k$ vertices, $e\stackrel{def}{=}x_ix_j$, and
$d(e)\stackrel{def}{=}d(x_i,x_j)$.
\end{definition}

In Fig.~\ref{diffdistance_fig}(b), we show the maps of transitive distance
matrices for the data set in Fig.~\ref{diffdistance_fig}(a) with orders from $1$
to $6$, where a larger intensity denotes a smaller transitive distance. In this
data set, there are 50 samples, and the samples in each cluster are
consecutively labeled.
From these maps, we can see that when $k$ is larger, the ratios of the
inter-cluster transitive distances to the intra-cluster transitive distances
tend to be larger. In other words, if more messengers are involved, the obtained
transitive distances better represent the relationship among the samples.

When the order $k=n$, where $n$ is the number of all the samples, we denote
$D_n$ with $D$ for simplicity. The following proposition shows that $D$
is an ultrametric.

\begin{proposition}\label{propultra}
The transitive distance $D$ is an ultrametric on a given data set.
\end{proposition}

The proof of Proposition~\ref{propultra} is simple and omitted here.
So given a data set $V$ and its distance matrix $E$, we can obtain another
ultrametric distance matrix $E'$ through Definition~\ref{transdk}.
In~\cite{ding2006tca}, an $O(n^3)$ algorithm is given to derive $E'$
from $E$. In Section \ref{secalgo}, we propose an algorithm which is almost
$O(n^2)$ to obtain $E'$.

It is worth mentioning that although we use $d(\cdot,\cdot)$ to denote the
Euclidean distance for convenience in the previous discussion, we can replace
$d(\cdot,\cdot)$ with any other traditional distance~(metric) in
Definition~\ref{transdk} and still have Proposition~\ref{propultra}. Therefore,
in what follows, $d(\cdot,\cdot)$ is used to denote any traditional distance. 

\subsection{Kernel Trick by the Transitive Distance}
\label{subkernel}
In this section, we show that the derived ultra-metric well reflects the
relationship among data samples and a kernel mapping with a promising property
can be
obtained. First we introduce a lemma from \cite{lemin1985iii} and
\cite{fiedler1998use}.

\begin{lemma}\label{embeddinglm}
Every finite ultrametric space consisting of $n$ distinct points can be
isometrically embedded into a $n-1$~dimensional Euclidean space.
\end{lemma}

With Lemma~\ref{embeddinglm}, we have the mapping\footnote{We use
$d(\cdot,\cdot)$ to denote a traditional distance in $V$ and $d'(\cdot,\cdot)$
the Euclidean distance in $V'$.}
\begin{equation}
\phi: (V\subset R^l,D) \rightarrow (V'\subset R^s,d'),
\end{equation}
where $\phi(x_i)=x'_i\in V'$, $s=n-1$, and $n$ is the number of points in a set
$V$. We also have $d'(\phi(x_i),\phi(x_j))=D(x_i,x_j)$, where $d'(\cdot,\cdot)$
is the Euclidean distance in $R^s$, i.e., the Euclidean distance between two
points in $V'$ equals its corresponding ultrametric distance in $V$.

Before giving an important theorem, we define the consistency stated in
Section~\ref{introsec} precisely.

\begin{definition}\label{consistency_def}
A labeling scheme $\{(x_i,l_i)\}$ of a data set $V=\{x_i | i=1,2,\cdots,n\}$,
where $l_i$ is the cluster label of $x_i$, is called consistent with some
distance $d(\cdot,\cdot)$ if the following condition holds: for any $y\notin C$
and any partition $C=C_1\cup C_2$, we have $d(C_1,C_2) < d(y,C)$, where
$C\subset V$ is
some cluster, $y\in V$, $d(C_1,C_2)\stackrel{def}{=}\min_{{x_i\in
C_1}\atop{x_j\in
C_2}}d(x_i,x_j)$ is the distance between the two sets $C_1$ and $C_2$, and
$d(y,C) \stackrel{def}{=} \min_{x\in C}d(y,x)$ is the distance between a point
$y$ and the set $C$.
\end{definition}

The consistency requres that the intra-cluster distance is strictly smaller
than the inter-cluster distance. This might be too strict in some practical
applications, but it helps us reveal the following desirable property
for clustering.

\begin{theorem}
If a labeling scheme of a data set $V=\{x_i | i=1,2,\cdots,n\}$, is consistent
with a distance $d(\cdot,\cdot)$, then given the derived transitive distance $D$
and the embedding $\phi:(V,D)\rightarrow (V',d')$, the convex hulls of the
images of the
clusters in $V'$ do not intersect with each other.
\label{convexity_thrm}
\end{theorem}

\begin{figure}
\centering
\includegraphics[width=9cm]{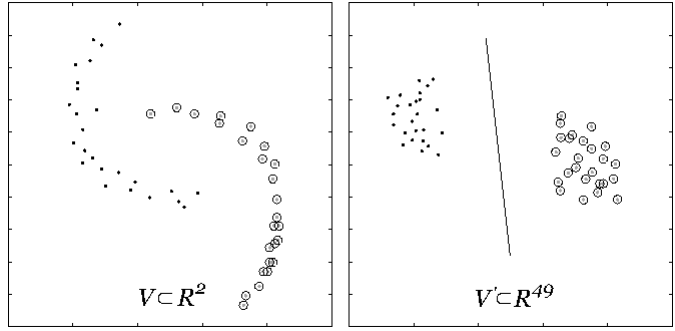}
\caption{Mapping a set of 50 data samples in $V\subset R^2$ to $V'\subset
R^{49}$.}
\label{figtheorem1}
\end{figure}

The proof of the theorem can be found in Appendix A.
An example of the theorem is illustrated in Fig.~\ref{figtheorem1}. A data set
$V$ with $50$ points in $R^2$ is mapped (embedded) into $R^{49}$, a much higher
dimensional Euclidean space, where the convex hulls of the two clusters do not
intersect. Moreover, the Euclidean distance between any two samples in $V'$
is equal to the transitive distance between these two samples in
$V$. The convex hulls of the two clusters intersect in $R^2$ but do not in
$R^{49}$, meaning that they are linearly separable in a higher dimensional
Euclidean space. We can see that the embedding $\phi$ is a desirable kernel
mapping.

Obviously, the clustering of $V'$ is much easier than the clustering of $V$. It
seems that the K-means algorithm can be used to perform the clustering of $V'$
easily.
Unfortunately, we only have the distance matrix $E'=[d'_{ij}]=[D_{ij}]$ of $V'$,
instead of the coordinates of $x'_i\in V'$, which are necessary for the K-means
algorithm. In Section~\ref{dualitysec}, we explain how to circumvent this
problem.

\section{K-Means Duality}
\label{dualitysec}
Let $E=[d_{ij}]$ be the distance matrix obtained from a data set
$V=\{x_i|i=1,2,\cdots,n\}$. From $E$, we can derive a new set
$Z=\{z_i|i=1,2,\cdots,n\}$, with $z_i\in R^n$ being the $i$th row of
$E$. Then we have the following observation, called the \emph{duality of
the K-means algorithm}.

\vspace{0.2cm}
\noindent\textbf{Observation}~(K-means duality):
\emph{The clustering result obtained by the K-means algorithm on $Z$ is
very similar to that obtained on $V$ if the clusters in $V$ are
hyperellipsoid-shaped}.
\vspace{0.2cm}

\begin{figure}
\centering
\includegraphics[width=9cm]{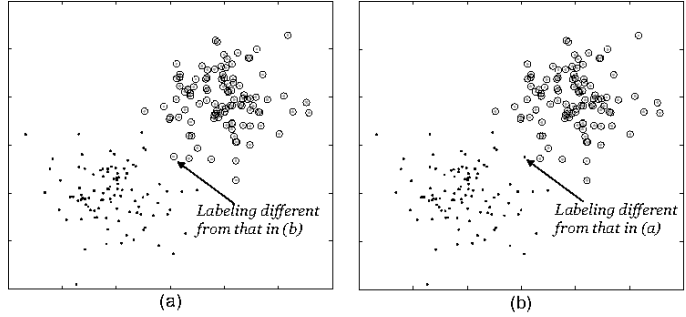}
\caption{(a) Clustering result obtained by the K-means algorithm on the original
data set $V$. (b) Clustering result obtained by the K-means algorithm on
$Z$ derived from the distance matrix of $V$. Only one sample has different
labelings from the two results.}
\label{dualityfig}
\end{figure}

We have this observation based on a large number of experiments on different
data sets. Most data sets were randomly generated with multi-Gaussian
distributions. From more than 100 data sets where each set contains 200
samples, we compared the results obtained
by the K-means alogrithms on original data sets $V$'s and their corresponding
sets $Z$'s. As a whole, the sample labeling difference is only 0.7\%. One
example is shown in Fig.~\ref{dualityfig}, in which only one sample is labeled
differently by the two clustering methods.

The matrix perturbation theory~\cite{stewart1990mpt} can be used to explain
this observation. We begin with an ideal case by supposing that the 
inter-cluster sample distances are much larger than the intra-cluster sample
distances (obviously, the clustering on this kind of data sets is easy). In the
ideal case, let the distance between any two samples in the same cluster be
$0$. If the samples are arranged in such a way that those in the same cluster
are indexed by successive integers, then the distance
matrix will be such a matrix:
\begin{equation}
    \hat{E} = \left(\begin{array}{cccc}
        E_1 & \cdots & \cdots & \cdots \\
        \cdots & E_2 & \cdots & \cdots \\
        \cdots & \cdots &\cdots&\cdots\\
        \cdots & \cdots & \cdots & E_k
        \end{array}\right)\begin{array}{cl}
    \} & n_1~rows \\
    \} & n_2~rows \\
    ~&~ \\
    \} & n_k~rows
    \end{array}
\end{equation} where $E_i=\mathbf{0}, 1\leq i\leq k$, represents the distance
matrix within the $i$th cluster, $n_1+n_2+\cdots+n_k=n$, and $k$ denotes the
number of clusters. Let $\hat{Z}=\{\hat{z}_i|i=1,2,\cdots,n\}$ with $\hat{z}_i$
being the $i$th row of $\hat{E}$. Then in this ideal case, we have
$\hat{z}_1=\hat{z}_2=\cdots=\hat{z}_{n_1},\hat{z}_{n_1+1}=\hat{z}_{n_1+2}
=\cdots=\hat{z}_{n_1+n_2},\cdots,\hat{z}_{n-n_k+1}=\hat{z}_{n-n_k+2}=\cdots=\hat
{z}_n$. Therefore, if $\hat{Z}$ is considered as a data
set to be clustered, the distance between any two samples in each cluster is
still $0$. On the other hand, for two samples in different clusters, say,
$\hat{z}_1$ and $\hat{z}_{n_1+1}$, we have
\begin{align}
\hat{z}_1 &= (\overbrace{0, \cdots,
0}^{n_1},d_{1,{n_1+1}},\cdots,d_{1,{n_1+n_2}},\cdots), \\
\hat{z}_{n_1+1} &= (d_{n_1+1,1},\cdots,d_{n_1+1,n_1},\underbrace{0, \cdots,
0}_{n_2},d_{n_1+1,n_1+n_2+1},\cdots),
\end{align}
and
\begin{equation}
d(\hat{z}_1,\hat{z}_{n_1+1}) \geq \sqrt{\sum_{j=n_1+1}^{n_1+n_2}d_{1,j}^2 +
\sum_{j=1}^{n_1}d_{n_1+1,j}^2}\gg 0.
\end{equation}

Thus, the distance between any two samples in different clusters is still
large. The distance relationship in the original data set is preserved
completely in this new data set $\hat{Z}$. Obviously, the K-means algorithm on
the original data set can give the same result as that on $\hat{Z}$ in this
ideal case. In general cases, a perturbation $P$ is added to $\hat{E}$, i.e.,
$E=\hat{E}+P$, where all the diagonal elements of $P$ are zero. The matrix
perturbation theory~\cite{stewart1990mpt} indicates that the K-means clustering
result on the data set $Z$ that is derived from $E$ is similar to that on
$\hat{Z}$ if $P$ is not dominant over $\hat{E}$. Our experiments and the above
analysis support the observation of the K-means duality.

Now we are able to give a solution to the problem mentioned at the end of
Section~\ref{subkernel}. From Theorem~\ref{convexity_thrm}, we can map a data
set $V$ to $V'\subset R^{n-1}$ where the clustering is easier if the
clusters with the original distance are consistent in $V$. The problem we need
to
handle is that in $R^{n-1}$ we only have the distance matrix instead of the
coordinates of the samples in $V'$. From the analysis of the K-means duality in
this section, we can perform the clustering based on the distance matrix by the
K-means algorithm. Therefore, the main ingredients for a new clustering
algorithm are
already available.

\section{A New Clustering Algorithm}
\label{secalgo}
Given a data set $V=\{x_i|i=1,2,\cdots,n\}$, our clustering algorithm is
described
as follows.

\begin{algorithm}\baselineskip=18pt
\caption{Clustering Based on the Transitive Distance and the K-means Duality}
\begin{itemize}
\item[1)] Construct a weighted complete graph $G=(V,E)$ where
$E=[d_{ij}]_{n\times n}$ is the distance matrix containing the weights of
all the edges and $d_{ij}$ is the distance between samples $x_i$ and $x_j$.
\item[2)] Compute the transitive distance matrix $E'=[d'_{ij}]=[D_{ij}]$ based
on
$G$ and Definition \ref{transdk}, where $D_{ij}$ is the transitive distance
with order $n$ between samples $x_i$ and $x_j$.
\item[3)] Perform clustering on the data set $Z'=\{z'_i|i=1,2,\cdots,n\}$ with
$z'_i$ being the $i$th row of $E'$ by the K-means algorithm and then assign the
cluster
label of $z'_i$ to $x_i$, $i=1,2,\cdots, n$.
\end{itemize}
\label{algomain}
\end{algorithm}

In step 2), we need to compute the transitive distance with order $n$ between
any two samples in $V$, or equivalently, to find the \emph{transitive edge},
which is defined below.

\noindent\begin{definition}
For a weighted complete graph $G=(V,E)$ and any two vertices $x_p,x_q\in V$, the
transitive edge for the pair $x_p$ and $x_q$ is an edge $e=x_ux_v$, such that
$e$ lies
on a path connecting $x_p$ and $x_q$ and $D_{pq}=D(x_p,x_q)=d(x_u,x_v)$.
\end{definition}

An example of a transitive edge is shown in Fig.~\ref{diffdistance_fig}(a).
Because the number of paths between two vertices (samples) is exponential in the
number of the samples, the brutal searching for the transitive distance between
two samples is infeasible. It is necessary to design a faster algorithm to carry
out this task. The following Theorem \ref{mst_thrm} is for this purpose.

Without loss of generality, we assume that the weights of edges in $G$ are
distinct. This can be achieved by slight perturbations of the positions of the
data samples. After this modification, the clustering result of the data will
not be changed if the perturbation are small enough.

\begin{theorem}
Given a weighted complete graph $G=(V,E)$ with distinct weights, each transitive
edge lies on the minimum spanning tree $\widetilde{G}=(V,\widetilde{E})$ of $G$.
\label{mst_thrm}
\end{theorem}

The proof of Theorem~\ref{mst_thrm} can be found in Appendix~B. This
theorem suggests an efficient algorithm to compute the transitive
matrix $E'=[d'_{ij}]_{n\times n}$ which is shown in
Algorithm~\ref{transitive_alg}. Next we analyze the computational
complexity of this algorithm.

\begin{algorithm}\baselineskip=12pt
   \caption{Computing the transitive distance matrix
   $E'=[d'_{ij}]_{n\times n}$}
    \begin{itemize}
    \item[1)] Build the minimum spanning tree
    $\widetilde{G}=(V,\widetilde{E})$ from $G=(V,E)$.
    \item[2)] Initialize a forest $F\leftarrow\widetilde{G}$.
    \item[3)] \textbf{Repeat}
    \item[4)] \hspace{0.4cm}\textbf{For} each tree $T\in F$ \textbf{do}
    \item[5)] \begin{itemize}
              \item[]Cut the edge with the largest weight $w_T$ and partition
$T$ into $T_1$ and $T_2$.
              \end{itemize}
    \item[6)] \hspace{0.7cm}\textbf{For} each pair $(x_i,x_j)$,
    $x_i\in T_1$, $x_j\in T_2$ \textbf{do}
    \item[7)] \hspace{1.1cm}$d'_{ij}\leftarrow w_T$
    \item[8)] \hspace{0.7cm}\textbf{End for}
    \item[9)] \hspace{0.4cm}\textbf{End for}
    \item[10)] \textbf{Until} each tree in $F$ has only one vertex.
    \end{itemize}
   \label{transitive_alg}
\end{algorithm}

Building the minimum spanning tree from a complete graph $G$ needs
time very close to $O(n^2)$ by the algorithm in
\cite{chazelle2000mst}\footnote{The fastest
algorithm~\cite{chazelle2000mst} to obtain a minimum spanning tree
needs $O(e\alpha(e,n))$ time, where $e$ is the number of edges and
$\alpha(e,n)$ is the inverse of the Ackermann function. The function
$\alpha$ increases extremely slowly with $e$ and $n$, and therefore
in practical applications it can be considered as a constant not
larger than $4$. In our case, $e=O(n^2)$ for a complete graph, so 
the complexity for building a minimum spanning tree is
about $O(n^2)$.}. When Algorithm~\ref{transitive_alg} stops, total
$n$ non-trivial tree\footnote{A non-trivial tree is a tree with at
least one edge.} have been generated. The number of the edges in
each non-trivial tree is not larger than $n$. Therefore, the total
time taken by searching for the edge with the largest weight on each
tree (step 5) in the algorithm is bounded by
$O(n^2)$. Steps 6--8 are for finding the values for the elements of
$E'$. Since each element of $E'$ is visited only once, the total
time consumed by steps 6--8 is $O(n^2)$. Thus the computational
complexity of Algorithm~\ref{transitive_alg} is about $O(n^2)$.

Considering the time $O(n^2)$ for building the distance matrix $E$,
and the fact that the complexity of the K-means algorithm\footnote{The time
complexity of the K-means algorithm is $O(npq)$, where $p$ and $q$ are the
number of iterations and the dimension of the data samples, respectively. The
data set $Z'$ in Algorithm~\ref{algomain} is in $R^n$ and thus $q=n$. 
In practical applications, $p$ can be considered as smaller than a fixed 
positive number.} is
close to $O(n^2)$, we conclude that the computational complexity of
Algorithm~\ref{algomain} is about $O(n^2)$.

\begin{figure}
\centering
\includegraphics[width=10cm]{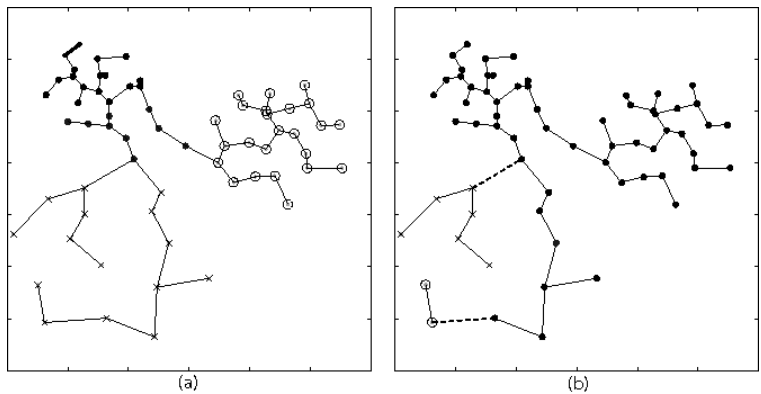}
\caption{(a) The minimum spanning tree and the clustering result by
our algorithm. (b) The minimum spanning tree and the clustering
result by the hierarchical clustering. The dashed lines are the cutting edges.
The number of clusters is 3.}\label{diff_hier_mg}
\end{figure}

Although the minimum spanning tree is used to help clustering in both the
hierarchical
clustering and our algorithm, the motivations and effects are quite different.
In
our case, the minimum spanning tree is for generating a kernel effect (to
obtain the relationship among the samples in a high dimensional space according
to
Theorem~\ref{convexity_thrm}), with which the K-means algorithm provides a
global
optimization function for clustering. Whereas in the hierarchical
clustering, each iteration step only focuses on the local sample distributions.
This difference leads to distinct algorithms in handling the data obtained from
the minimum spanning tree. We carry out the K-means algorithm on the derived
$Z'$
according to the K-means duality, while the hierarchical clustering cuts $c-1$
largest edges from the minimum spanning tree, where $c$ is the number of
clusters. In
Fig.~\ref{diff_hier_mg}, we show a data set clustered by the two approaches.
The multi-scale data set makes the hierarchical clustering give an unreasonable
result.

\section{Experiments}
\label{experiments_sec} We have applied the proposed algorithm to a number of
clustering problems to test its performance. The results are compared with those
by the K-means algorithm, the NJW spectral clustering algorithm~\cite{ng2001sca}
and the self-tuning spectral clustering algorithm~\cite{zelnikmanor2004sts}.
For each data set, the NJW algorithm needs manually tuning of the scale and the
self-tuning algorithm needs to set the number of nearest neighbors. On the
contrary, no parameters are required to set for our algorithm. In this
comparisons, we show the best clustering results that are obtain by adjusting
the parameters in the two spectral clustering algorithms. All the numbers of
clusters are assumed to be known.

\begin{figure*}
    \includegraphics[width=17cm]{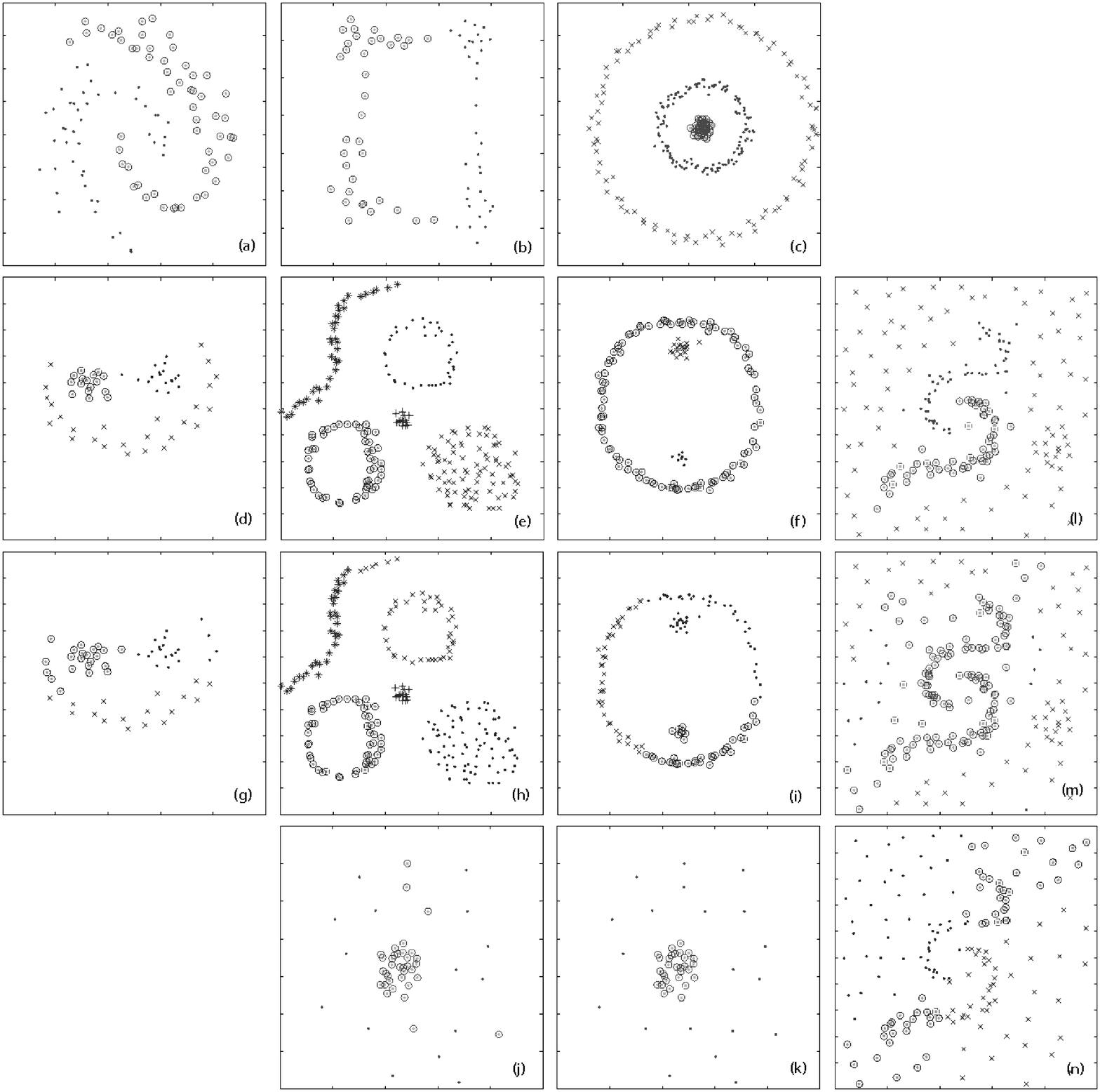}
    \caption{Clustering results by our algorithm and the two spectral
algorithms. (a)(b)(c) Results by the three algorithms. (d)(e)(f)
    Results by the NJW algorithm and ours. (g)(h)(i) Results by the self-tuning
    algorithm. (j) Result by the NJW algorithm. (k) Result by the self-tuning
    algorithm and ours. (l)(m)(n) Results by our algorithm, the NJW algorithm,
and the self-tuning algorithm, respectively.}
    \label{mgraph_synthetic_fig}
\end{figure*}

\subsection{Synthetic Data Sets}
Eight synthetic data sets are used in the experiments. Bounded in a region
$(0,1)\times (0,1)$, these data sets are with complex cluster shapes,
multi-scale clusters, and noise. The clustering results are shown in
Fig.~\ref{mgraph_synthetic_fig}. Note that the results obtained by the K-means
algorithm are not given because it is obvious that it cannot deal with these
data sets.

In Figs.~\ref{mgraph_synthetic_fig}(a)--(c), all the three algorithms obtain the
same results. Figs.~\ref{mgraph_synthetic_fig}(d)--(f) and (g)--(i) show three
data sets
on which the self-tuning algorithm gives different results from the other two
algorithms. The self-tuning algorithm fails to cluster the data sets no matter
how we tune its parameter. Figs.~\ref{mgraph_synthetic_fig}(j) and (k) show
two clustering results where the data set is with multi-scale clusters. The
former is
produced by the NJW algorithm and the latter by the self-tuning and our
algorithms. To cluster the data set in Figs.~\ref{mgraph_synthetic_fig}(l)--(n)
is a
challenging task, where two relatively tightly connected clusters are surrounded
by uniformly distributed noise samples (the third cluster). Our algorithm
obtains the more reasonable result~(Fig.~\ref{mgraph_synthetic_fig}(l)) than the
results by another two algorithms~(Figs.~\ref{mgraph_synthetic_fig}(m) and
(n)). 

From these samples, we can see that our algorithm performs similar to or better
than the NJW and self-tuning spectral clustering algorithms. This statement
applies to many other data sets we have tried, which are not shown here due to
the limitation of space.

\subsection{Data Sets from the USPS Database}
USPS database is an image database provided by the US Postal Service. There are
9298 handwriting digit images of size $16\times 16$ from ``0'' to ``9'' in the
database, from which we construct ten data sets from this database. Each set has
1000 images selected randomly with two, three, or four clusters. Each image is
treated as a point in a 256-dimensional Euclidean space. The following figure
shows the error rates of the four algorithms on these sets. In this experiments,
the parameters for the NJW and self-tuning algorithms are tuned carefully to
obtain the smallest error rates. These results show that as a whole, our
algorithm achieves the smallest error rate, and the K-means and self-tuning
algorithms perform worst.
\begin{figure}
\centering \includegraphics[width=11cm]{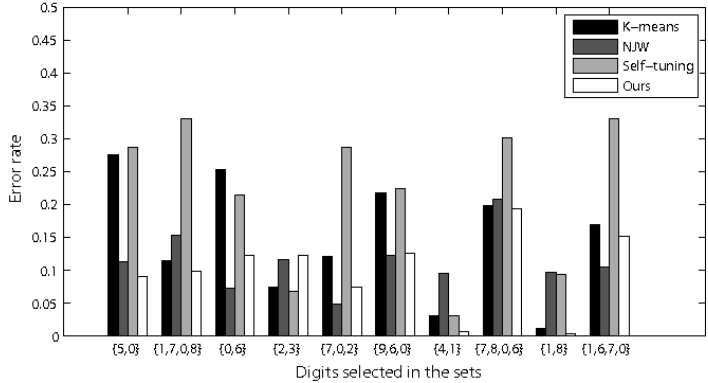} \caption{The error rates
of the four algorithms on the ten data sets constructed from the USPS database.}
\label{usps_fig}
\end{figure}

\subsection{Iris and Ionosphere Data Sets}
We also test the algorithms on two commonly-used data sets, Iris and Ionosphere,
in UCI machine learning database. Iris consists of 150 samples in 3 classes,
each with 50 samples. Each sample has 4 features. Ionosphere contains 354
samples in 2 classes and each sample has 34 features. In Table~\ref{uci_tab} we
show the error rates of the four algorithms clustering on these data sets. For
the NJW and self-tuning algorithms, we have to adjust their parameters~($\delta$
and $N$)\footnote{We tried different $\delta$ from $0.01$ to $0.1$ with step
$0.001$ and $0.1$ to $4$ with step $0.1$, and different $N$ from $2$ to $30$
with step $1$.} to obtain the smallest error rates, which are shown in the
table. Our algorithm results in the smallest error rates among the four
algorithms.

\begin{table}[h]
\centering \caption{Error rates of the four algorithms on Iris and
Ionosphere data sets}\label{uci_tab}
\begin{tabular}{|l|c|c|c|c|}
\hline
~ & K-means& NJW & Self-tuning & Ours \\
\hline Iris & 0.11 & 0.09 ($\delta=0.40$)
& 0.15 ($N=5$) & 0.07 \\
Ionosphere & 0.29 & 0.27 ($\delta=0.20$)
& 0.30 ($N=6$) & 0.15\\
\hline
\end{tabular}
\end{table}

\subsection{Remarks}
From the experiments, we can see that compared with the K-means algorithm, our
algorithm and the spectral algorithms can handle the clustering of a data set
with complex cluster shapes. Compared with the spectral algorithms, our
algorithm has comparable or better performance and does not need to adjust any
parameter. In the above experiments, since we have the ground truth for each
data set, we can try different parameters in the NJW and self-tuning algorithms
so that they produce the best results. However, we do not know which parameters
should be the best for unsupervised data clustering in many applications.
Another advantage of our algorithm over the spectral algorithms is that its
computational complexity is close to $O(n^2)$, while the spectral algorithms'
complexities are $O(n^3)$. 

\section{Conclusion}
\label{conclusion_sec} In this paper, we have built a connection between
the transitive distance and the kernel technique for data clustering, By using
the transitive distance, we show that if the consistency conditions is
satisfied, the clusters of arbitrary shapes can be mapped to a new space where
the clusters are easier to be seperated. Based on the observed K-means duality,
we have developed an efficient algorithm with computational complexity $O(n^2)$.
Compared with the two popular spectral algorithms whose computational
complexities are $O(n^3)$, our algorithm is faster, without the need to tune any
parameters, and performs very well. Our algorithm can be used to handle
challenging clustering problems where the data sets are with complex shapes,
multi-scale clusters, and noise.

\section{Appendix A: Proof of Theorem 1}
\label{secappdx}
It is reasonable to assume that each cluster has at least two samples. Let
$x_i$, $x_j\in C$, $x_k\notin C$, $x_i$, $x_j$, $x_k\in V$, where $C\subset V$
is some cluster. Then their images after the mapping $\phi$ are $x'_i$, $x'_j$,
$x'_k\in V'$, where $x'_i$, $x'_j\in C'$, $x'_k\notin C'$, and $C'=\phi(C)$.
\begin{itemize}
    \item[(i)] First, we verify that if
    $d'(x'_i,x'_j)\geqslant d_0\in R^+$, then there exists a partition $C_1\cup
C_2=C$ such that $d(C_1,C_2)\geqslant d_0$.
    Such a partition can be obtained by the following steps:
    \begin{itemize}
        \item[1)] Initialize $H=C$, $m=1$, $C_1=\emptyset$, and
        $C_2=\emptyset$.
        \item[2)] Find a path $\mathcal{P}$ including the transitive edge
        from $x_i$ to $x_j$ in $H$.
        \item[3)] Cut the transitive edge on the path $\mathcal{P}$. Let
        $\mathcal{P}_m$ ($\mathcal{Q}_m$) be the set consisting of the
        samples on $\mathcal{P}$ that are on the same side with $x_i$
        ($x_j$) after the cutting, except $x_i$ ($x_j$).
        \item[4)] $C_1\leftarrow C_1\cup \mathcal{P}_m$,
        $C_2\leftarrow C_2\cup \mathcal{Q}_m$,
        $H\leftarrow H\backslash\{\mathcal{P}_m\cup\mathcal{Q}_m\}$,
        and $m\leftarrow m+1$.
        \item[5)] Repeat 2), 3), and 4) until only $x_i$ and $x_j$
        are left in $H$.
        \item[6)] $\mathcal{P}_m\leftarrow \{x_i\}$,
        $\mathcal{Q}_m\leftarrow \{x_j\}$,
        $C_1\leftarrow C_1\cup\mathcal{P}_m$, and
        $C_2\leftarrow C_2\cup\mathcal{Q}_m$.
    \end{itemize}
    In this procedure, from (\ref{transdis_equ}) we can see that
        $d(\mathcal{P}_s,\mathcal{Q}_t)\geqslant d'(x'_i,x'_j),
        ~1\leqslant s,t\leqslant m$.
    Since
        $C_1 = \mathcal{P}_1\cup\mathcal{P}_2\cup\cdots\cup\mathcal{P}_m$
    and
        $C_2 = \mathcal{Q}_1\cup\mathcal{Q}_2\cup\cdots\cup\mathcal{Q}_m$,
    we have
        $d(C_1,C_2) = \min_{1\leqslant s,t\leqslant m}
        \{d(\mathcal{P}_s,\mathcal{Q}_t)\}$.
    Thus,
        $d(C_1,C_2)\geqslant d'(x'_i,x'_j)\geqslant d_0$.
    \item[(ii)] Second, we show that there exist
    $x_u\in C$ and $x_v\notin C$ such that
    $d'(x'_i,x'_k)\geqslant d(x_u,x_v)$.
    From Definition
    \ref{transdk}, we have a path $\mathcal{P}$ connecting
    $x_i$ and $x_k$ including the transitive edge.
    Then there exists an edge
    $x_ux_v\in\mathcal{P}$ such that $x_u\in C$ and $x_v\notin C$,
    and from (\ref{transdis_equ}), we have
    $d'(x'_i,x'_k)\geqslant d(x_u,x_v)$.
    \item[(iii)] Third, we show that
    \begin{equation}
        d'(x'_i,x'_j)\leqslant\min\{d'(x'_i,x'_k),d'(x'_j,x'_k)\}.
    \label{kernelinv_equ}
    \end{equation}
    Assume, to the contrary, that
        $d'(x'_i,x'_j)>d'(x'_i,x'_k)$.
    From (i) and (ii), we have a partition $C_1\cup C_2=C$, and
    $x_u\in C$, $x_v\notin C$ such that
    $d(C_1,C_2)\geqslant d'(x'_i,x'_j)$ and
    $d'(x'_i,x'_k)\geqslant d(x_u,x_v)$. Thus
        $d(C_1,C_2)\geqslant d'(x'_i,x'_j)>d'(x'_i,x'_k)
        \geqslant d(x_u,x_v)\geqslant d(C,x_v)$,
    which contradicts the consistency of $V$. Therefore,
    (\ref{kernelinv_equ}) holds.
    \item[(iv)] Let
        $C=\{x_{c_1},\cdots,x_{c_m}\}$
    be a cluster in $V$, with its image
        $C'=\phi(C)=\{x'_{c_1},\cdots,x'_{c_m}\}\subset V'$.
    Let $\widetilde{C}'$ be the convex hull of $C'$. Now we verify that
    no samples not in $C'$ are in $\widetilde{C}'$. Assume, to the
    contrary, that there exists a
    sample $y'\in\widetilde{C}'$, $y\notin C'$. Consider
    a sample $x'\in C'$. Let $P$ be the hyperplane, each point on which
    has the same distance to $x'$ and
    $z'$. Then there must exist another sample
    $z'\in C'$ such that $y'$ and $z'$ are in the same side of $P$,
    which leads to $d'(x',z')>d'(y',z')$, a contradiction to
    (\ref{kernelinv_equ}).
\end{itemize}
In (iv), we have verified that for any cluster $C'\in V'$, no
samples from other clusters can be in the convex hull of $C'$. Thus,
the convex hulls of all the clusters in $V'$ are not intersecting
each other.

\section{Appendix B: Proof of Theorem 2}
For any two distinct vertices $x_1$ and $x_2$ in $G$, let
$\mathcal{P}=x_{k_1}x_{k_2}\cdots x_{k_s}$ be the path connecting
them including the transitive edge $x_{k_i}x_{k_{i+1}}$, where $k_1=1$
and $k_s=2$. Then from Definition \ref{transdk}, we have
\begin{equation}
 d(x_{k_m},x_{k_{m+1}}) < d(x_{k_i},x_{k_{i+1}}),~
m=1,2,\cdots,i-1,i+1,\cdots,s.
\end{equation}
Next we verify that the edge $x_{k_i}x_{k_{i+1}}$ is in $\widetilde{G}$.
Let $\widetilde{G}_{\mathcal{P}}=\widetilde{G}\cup\mathcal{P}$.
Assume, to the contrary, that
$x_{k_i}x_{k_{i+1}}\notin\widetilde{G}$. Then the edge $x_{k_i}x_{k_{i+1}}$
must be on a loop $\mathcal{O}\subseteq
\widetilde{G}_{\mathcal{P}}$. Consider the following two cases:
\begin{itemize}
\item[(i)] For any edge $x_ux_v\in \widetilde{G}\cap\mathcal{O}$,
$d(x_u,x_v)<d(x_{k_i},x_{k_{i+1}})$.
\item[(ii)] There exists an edge $x_{l_j}x_{l_{j+1}}\in
\widetilde{G}\cap\mathcal{O}$
such that $d(x_{l_j},x_{l_{j+1}})>d(x_{k_i},x_{k_{i+1}})$.
\end{itemize}

Suppose that case (i) is true. Then for any edge on the path
$(\mathcal{P}\cup\mathcal{O})\backslash\{x_{k_i}x_{k_{i+1}}\}$ that
also connects $x_1$ and $x_2$, we have its length smaller than
the transitive edge for $x_1$ and $x_2$. Thus case (i)
cannot be true.

Suppose that case (ii) is true. Since
    $\widetilde{G}^*=(\widetilde{G}\cup\{x_{k_i}x_{k_{i+1}}\})
    \backslash\{x_{l_j}x_{l_{j+1}}\}$
is a spanning tree of $G$, and the sum of the edge weights in
$\widetilde{G}^*$ is smaller than that in $\widetilde{G}$, we have a
contradiction to the fact that $\widetilde{G}$ is the minimum
spanning tree. Thus case (ii) cannot be true either, which completes
the proof.

{\small
\bibliographystyle{plain}
\bibliography{ultrametric_duality-arxiv}
}
\end{document}